\documentclass[runningheads]{llncs}

\usepackage{eccv}

\usepackage{eccvabbrv}

\usepackage{graphicx}
\usepackage{booktabs}
\usepackage{subcaption} 

\usepackage[accsupp]{axessibility}  

\newlength\savewidth

\usepackage[dvipsnames]{xcolor}
\usepackage{colortbl}
\definecolor{baselinecolor}{gray}{.9}
\newcommand{\baseline}[1]{\cellcolor{baselinecolor}{#1}}
\newcommand{\mysection}[1]{\paragraph{\textbf{#1}}}

\usepackage{hyperref}

\usepackage{orcidlink}

\begin{document}

\title{Efficient Image Pre-Training with\\Siamese Cropped Masked Autoencoders} 


\titlerunning{CropMAE}

\author{Alexandre Eyma\"el$^*$\inst{1}\orcidlink{0009-0004-5305-7307} \and
Renaud Vandeghen$^*$\inst{1}\orcidlink{0009-0003-1752-1195} \and
Anthony Cioppa\inst{1,2}\orcidlink{0000-0002-5314-9015} \and \\
Silvio Giancola\inst{2}\orcidlink{0000-0002-3937-9834} \and
Bernard Ghanem\inst{2}\orcidlink{0000-0002-5534-587X} \and
Marc Van Droogenbroeck\inst{1}\orcidlink{0000-0001-6260-6487}}

\authorrunning{A.~Eyma\"el et al.}

\institute{University of Liège, Belgium \and
KAUST, Saudi Arabia \\
\email{r.vandeghen@uliege.be}}

\maketitle

\begin{abstract}
Self-supervised pre-training of image encoders is omnipresent in the literature, particularly following the introduction of Masked autoencoders (MAE). 
Current efforts attempt to learn object-centric representations from motion in videos.
In particular, SiamMAE recently introduced a Siamese network, training a shared-weight encoder from two frames of a video with a high asymmetric masking ratio (95\%).
In this work, we propose CropMAE, an alternative approach to the Siamese pre-training introduced by SiamMAE.
Our method specifically differs by exclusively considering pairs of cropped images sourced from the same image but cropped differently, deviating from the conventional pairs of frames extracted from a video.
CropMAE therefore alleviates the need for video datasets, while maintaining competitive performances and drastically reducing pre-training and learning time. Furthermore, we demonstrate that CropMAE learns similar object-centric representations without explicit motion, showing that current self-supervised learning methods do not learn such representations from explicit object motion, but rather thanks to the implicit image transformations that occur between the two views.
Finally, CropMAE achieves the highest masking ratio to date (98.5\%), enabling the reconstruction of images using only two visible patches. Our code is available at \url{https://github.com/alexandre-eymael/CropMAE}.
\end{abstract}

\section{Introduction}
\label{sec:Intro}

Self-supervised learning (SSL) has become increasingly popular in the last few years thanks to its capacity to learn meaningful and robust representation without the need for labels, sometimes even leading to performances on downstream tasks surpassing its supervised counterpart. 
This is especially interesting in domains in which data labelling is costly, such as image segmentation or object detection, or when the exact task to solve is not known beforehand~\cite{Balestriero2023ACookbook-arxiv}. 
Among popular self-supervised paradigms, visual contrastive learning~\cite{Chen2020Simple, He2020Momentum, Grill2020Bootstrap} and masked image modeling (MIM)~\cite{Kenton2019Bert, He2022Masked, Xie2021AnEmpirical} have received significant interest due to their impressive performance. 
While highly effective, MIM methods often require a large amount of data and/or extensive training time to achieve satisfactory performance~\cite{Gupta2023Siamese, Tong2022VideoMAE, Feichtenhofer2022MAEST}. 
This necessity largely stems from their objective to develop a conceptual understanding of the data distribution they are trained on, enabling them to reconstruct images at the pixel level. 
This challenge is particularly pronounced with Vision Transformers (ViTs)~\cite{Dosovitskiy2021ViT} as encoders, as they perform sub-optimally with limited data due to the lack of visual inductive biases that they exhibit~\cite{Dosovitskiy2021ViT}.
A major weakness of contrastive learning techniques is that they rely on carefully chosen transformations to achieve good performances~\cite{Xiao2021What, Chen2020Simple, Grill2020Bootstrap}.

\begin{figure}[t!]
    \centering
    \includegraphics[width=\columnwidth]{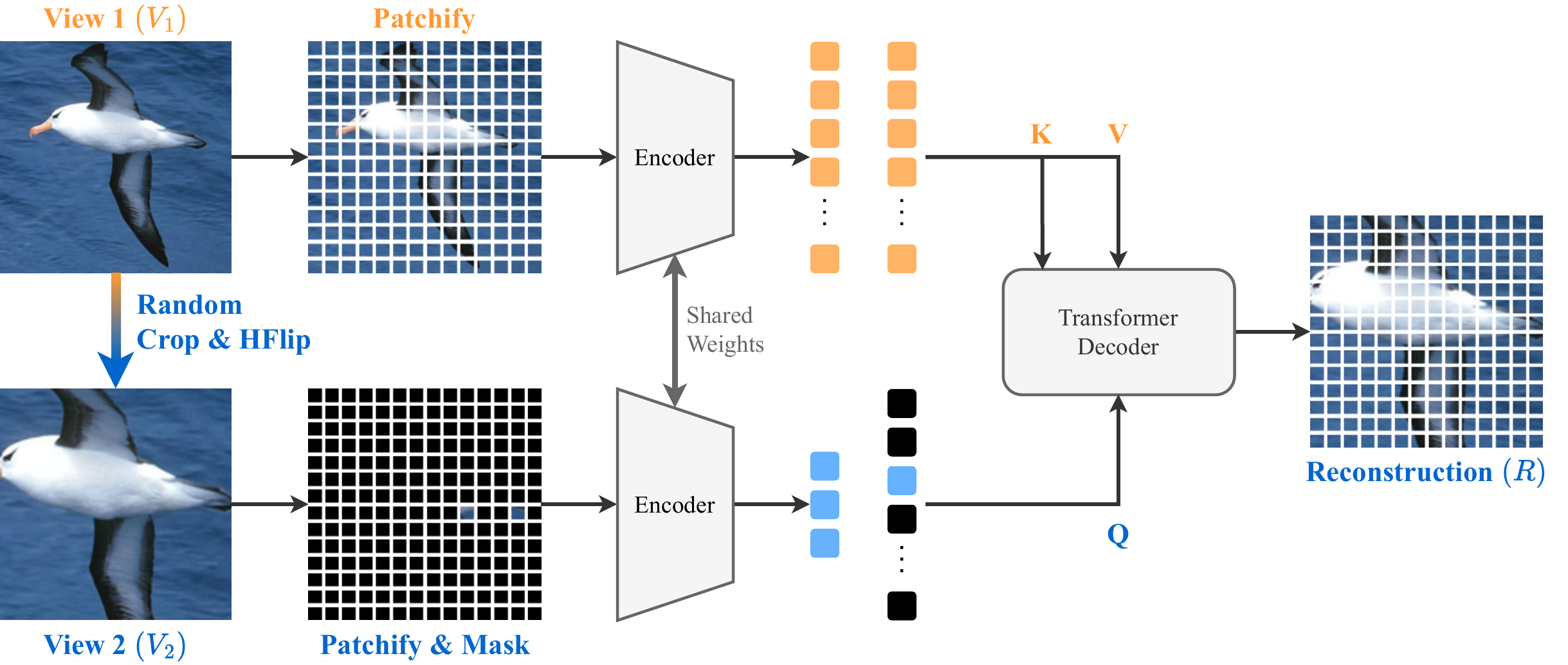}
    \caption{
    \textbf{CropMAE self-supervised pre-training.} 
    Given an input image ($V_1$), a second image is generated by performing a random crop and, optionally, a horizontal flip on the original image ($V_2$). 
    We then patchify~\cite{Dosovitskiy2021ViT} both views and mask~\cite{Kenton2019Bert, He2022Masked} an extremely high portion of the second image (above 98.5\%). 
    Both views are encoded by a Siamese~\cite{Bromley1993Siamese} ViT encoder, with added positional embedding~\cite{Dosovitskiy2021ViT}. 
    A transformer~\cite{Girdhar2017Attention} decoder reconstructs the masked image $R$ using self-attention layers on the tokens of the masked image and cross-attention layers between the tokens of the masked and unmasked images.
    }
    \label{fig:cropmae_pipeline}
\end{figure}

Recently, Siamese Masked autoencoders (SiamMAE)~\cite{Gupta2023Siamese} achieved state-of-the-art performance in numerous propagation tasks~\cite{Zhou2018Adaptive, PontTuset2017Davis-arxiv, Jhuang2013Towards} by learning object-centric representations from videos. 
This method leverages a Siamese encoder~\cite{Bromley1993Siamese} to process pairs of frames that are asymmetrically masked. 
Despite its impressive performance, SiamMAE faces two main limitations.
Firstly, it is designed to only process video frames, not standalone images. 
Yet, image datasets are typically orders of magnitude larger than video datasets, and less computationally expensive to decode, making image-based pre-training more effective and scalable than video-based pre-training.
Secondly, while SiamMAE reduces the need for the intense data augmentation found in contrastive learning methods, it still requires learning a conceptual understanding of the visual world, similar to most MIM techniques, thus requiring extensive training ($2{,}000$ epochs) on large datasets such as K400~\cite{Kay2017TheKinetics-arxiv} to reach state-of-the-art performances.

In this work, we propose a novel self-supervised learning method, called \emph{CropMAE}, that reframes the siamese-based paradigm introduced in SiamMAE in order to alleviate the need for video dataset, while keeping competitive performances on downstream tasks.
Specifically, we use random views of the same image to simulate viewpoint changes, object transformations, motion, and occlusions. 
Our method can therefore leverage both image and video datasets, and train at a significantly faster pace than SiamMAE. 
Moreover, we demonstrate that CropMAE learns meaningful object-centric representations for downstream video tasks without explicit motion.
Finally, unlike most MIM techniques, the pretext task of CropMAE is directly tractable based on the visible frame without the need to learn conceptual information about the world, which we believe is the reason for its faster training. An overview of our method is presented in~\Cref{fig:cropmae_pipeline}.

\mysection{Contributions.} We summarize our contributions as follows.
\textbf{(i)} We introduce a novel pre-training method, CropMAE, based on sole images, which alleviates the need for video decoding and significantly accelerates training.
The novel pretext task we introduce learns faster while quickly reaching good performances.
\textbf{(ii)} We empirically demonstrate the feasibility of learning meaningful representations for downstream video tasks from still images or data distributions traditionally not associated with videos. Notably, this approach yields better results than training directly on video frames.
\textbf{(iii)} We show, for the first time, that employing an extremely high masking ratio (98.5\%, \ie, using only two visible patches for a ViT/16), surpassing those explored in existing studies, can be optimal and generate a meaningful and challenging self-supervised task.

\section{Related Work}
\label{sec:SOTA}

\mysection{Visual representation learning.} Visual self-supervised learning focuses on learning rich and generalizable representations of images or videos. This is typically achieved through pretext tasks~\cite{Misra2016Shuffle, Noroozi2016Unsupervised, Gidaris2018Unsupervised, Chen2020Simple}, enabling the learned representations to be applicable to a broad set of downstream tasks~\cite{Everingham2010PascalVOC, Deng2009ImageNet, Zhou2018Adaptive}, either by fine-tuning the learned models for specific tasks, or by freezing the weights and training a linear classifier or an MLP on top of it. Key downstream tasks in the visual domain include image classification~\cite{Chen2020Simple, Grill2020Bootstrap, Caron2021Emerging, Chen2021Exploring, Xie2021AnEmpirical, zhou2021ibot, Bao2022BEiT, He2022Masked, Girdhar2023OmniMAE, Oquab2024DinoV2}, video classification~\cite{Tong2022VideoMAE, Wang2023VideoMAEV2, Feichtenhofer2022MAEST, Girdhar2023OmniMAE, Fan2023MotionGuided, Oquab2024DinoV2}, object detection~\cite{Grill2020Bootstrap, Chen2021Exploring, He2022Masked}, and video segmentation~\cite{Caron2021Emerging, Chen2021Exploring, Bao2022Discovering, Gupta2023Siamese, Jiang2023Concatenated-arxiv}. Our method, CropMAE, is a new visual self-supervised representation learning method for propagation tasks~\cite{PontTuset2017Davis-arxiv, Zhou2018Adaptive, Jhuang2013Towards}.

\mysection{Contrastive Self-Supervised Learning.} Contrastive self-supervised learning~\cite{Hadsell2006Dimensionality} has been recognized as an effective method for feature extraction, applicable both to images~\cite{Chen2021Exploring, Grill2020Bootstrap} and videos~\cite{Sermanet2018TimeContrastive, Dave2022TCLR}. This approach encourages the encoder to learn robust representations of the input data by minimizing the distance between representations of different augmented versions of the same image. Initially, it was common to enforce distinct images to have different representations in order to avoid representation collapse~\cite{Doersch2017Multitask, Wu2018Unsupervised, Chen2020Simple}. However, subsequent discoveries~\cite{He2020Momentum, Grill2020Bootstrap} have shown that robust learning can be achieved even without imposing this constraint. Contrastive self-supervised learning has also been widely used for correspondence learning~\cite{Wang2019Learning, Jabri2020SpaceTime}, as it inherently learns to build representations that are invariant and robust to perturbations. Contrary to contrastive learning, CropMAE does not rely as extensively on data augmentations and is not subject to representation collapse issues.

\mysection{Masked Image Modeling.} Drawing inspiration from the field of natural language processing~\cite{Kenton2019Bert}, masked image modeling (MIM) techniques have emerged as highly effective learners in the vision domain~\cite{Bao2022BEiT, He2022Masked, Xie2022SimMIM}. This approach involves dividing images into small patches~\cite{Dosovitskiy2021ViT}, with a high proportion of them being masked, and subsequently reconstructing them using a denoising autoencoder~\cite{Vincent2008Extracting}. Notably, after the training phase, the decoder is discarded, leaving the encoder to serve as a feature extractor. MIM has been applied with success across a broad range of fields, and has had numerous extensions and improvements~\cite{Tong2022VideoMAE, Wang2023VideoMAEV2, Bandara2023AdaMAE, Gupta2023Siamese, Feng2023Evolved, Qing2023MAR, Girdhar2023OmniMAE, Fan2023MotionGuided, Feichtenhofer2022MAEST, Jiang2023Concatenated-arxiv, Chen2023Context, Oquab2024DinoV2}. 

\mysection{Siamese Masked Autoencoders.} Building upon the work of masked autoencoders~\cite{He2022Masked}, Siamese Masked Autoencoders (SiamMAE)~\cite{Gupta2023Siamese} have emerged as a new state-of-the-art in video propagation tasks such as video object segmentation~\cite{PontTuset2017Davis-arxiv}, pose keypoint propagation~\cite{Jhuang2013Towards}, and semantic part propagation~\cite{Zhou2018Adaptive}. SiameseMAE uses a Siamese encoder~\cite{Bromley1993Siamese} to process either pairs~\cite{Gupta2023Siamese} or groups~\cite{Jiang2023Concatenated-arxiv} of frames, randomly selected from a video. A key feature of SiameseMAE is its asymmetric masking technique: the initial frame undergoes no masking, thereby serving as a complete reference, while a substantial portion (up to 95\%) of the second frame is masked. This setup encourages the network to accurately reconstruct the masked subsequent frames using the fully visible initial frame as a reference. The efficacy of SiameseMAE is believed to stem from its ability to effectively model object motion from videos and visual correspondence, learning the ``propagation'' and boundaries of objects from their observed positions in the past to their future locations, based on the few visible patches~\cite{Gupta2023Siamese}. In this work, we show that explicit motion derived from videos is not mandatory for Siamese masked autoencoders to learn object-centric representations. Particularly, we demonstrate that the ability to recognize object boundaries and acquire propagation skills can be effectively learned from still images.

\section{Method}
\label{sec:Method}

We propose a novel self-supervised method, namely \emph{CropMAE}, capable of learning valuable representations both from images and video frames.
First, we create two augmented views ($V_1$ and $V_2$) of an input image ($I$) by randomly cropping, resizing and horizontally flipping the original image (Sec.~\ref{subsec:cropping}).
Second, we patchify~\cite{Dosovitskiy2021ViT} both views $V_1$ and $V_2$ (Sec.~\ref{subsec:patching}) and mask~\cite{Kenton2019Bert,He2022Masked} an extremely high portion of the second view ($V_2$) (Sec.~\ref{subsec:masking}). 
Both views are encoded in a Siamese~\cite{Bromley1993Siamese} ViT encoder, with an additional positional embedding~\cite{Dosovitskiy2021ViT}. 
Third, a transformer~\cite{Girdhar2017Attention} decoder reconstructs a target image $R$ (Sec.~\ref{subsec:architecture}). 
The Siamese network and the decoder are trained by minimizing the L2 norm between the target $V_2$ and the reconstructed image $R$.
After such pre-training, the decoder is discarded, and we use the encoder as a feature extractor on downstream tasks. 
With this setup, we demonstrate that meaningful data augmentations, particularly random crops, can generate rich and useful object-centric representations for propagation tasks \emph{without} explicit motion.
~\Cref{fig:cropmae_pipeline} illustrates the main components of our method.

\subsection{Cropping}
\label{subsec:cropping}
Random crops have been widely used in visual self-supervised learning, especially in contrastive learning, where they are essential to reach excellent performances and develop robust representations~\cite{Chen2020Simple, Grill2020Bootstrap, Chen2021Exploring}.
Specifically, we examine four strategies inspired by the contrastive learning literature~\cite{Chen2020Simple}.

\begin{figure}[t!]
  \centering
  \begin{subfigure}[t]{0.24\textwidth}
    \includegraphics[width=\textwidth]{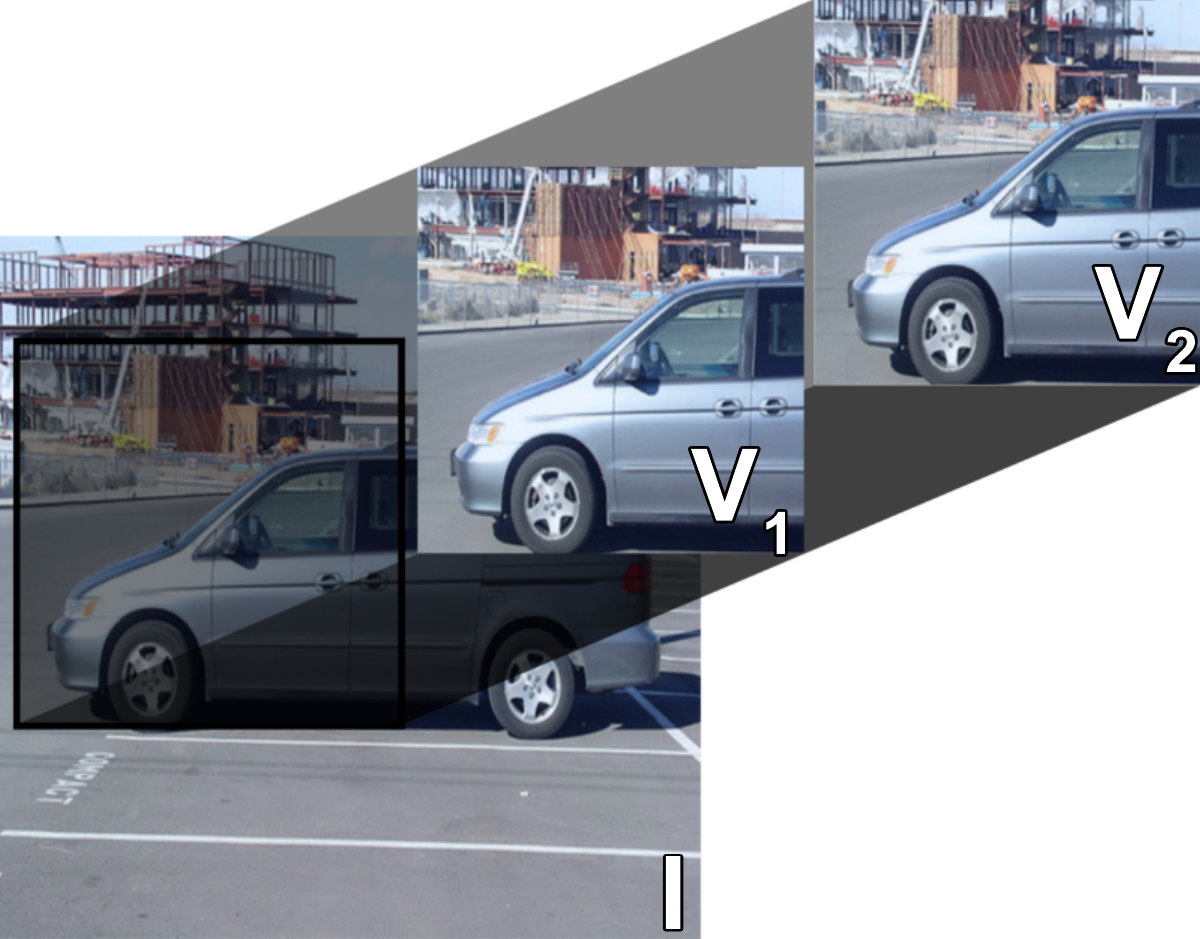}
    \caption{Same.}
    \label{subfig:same_views}
  \end{subfigure}
  \begin{subfigure}[t]{0.24\textwidth}
    \includegraphics[width=\textwidth]{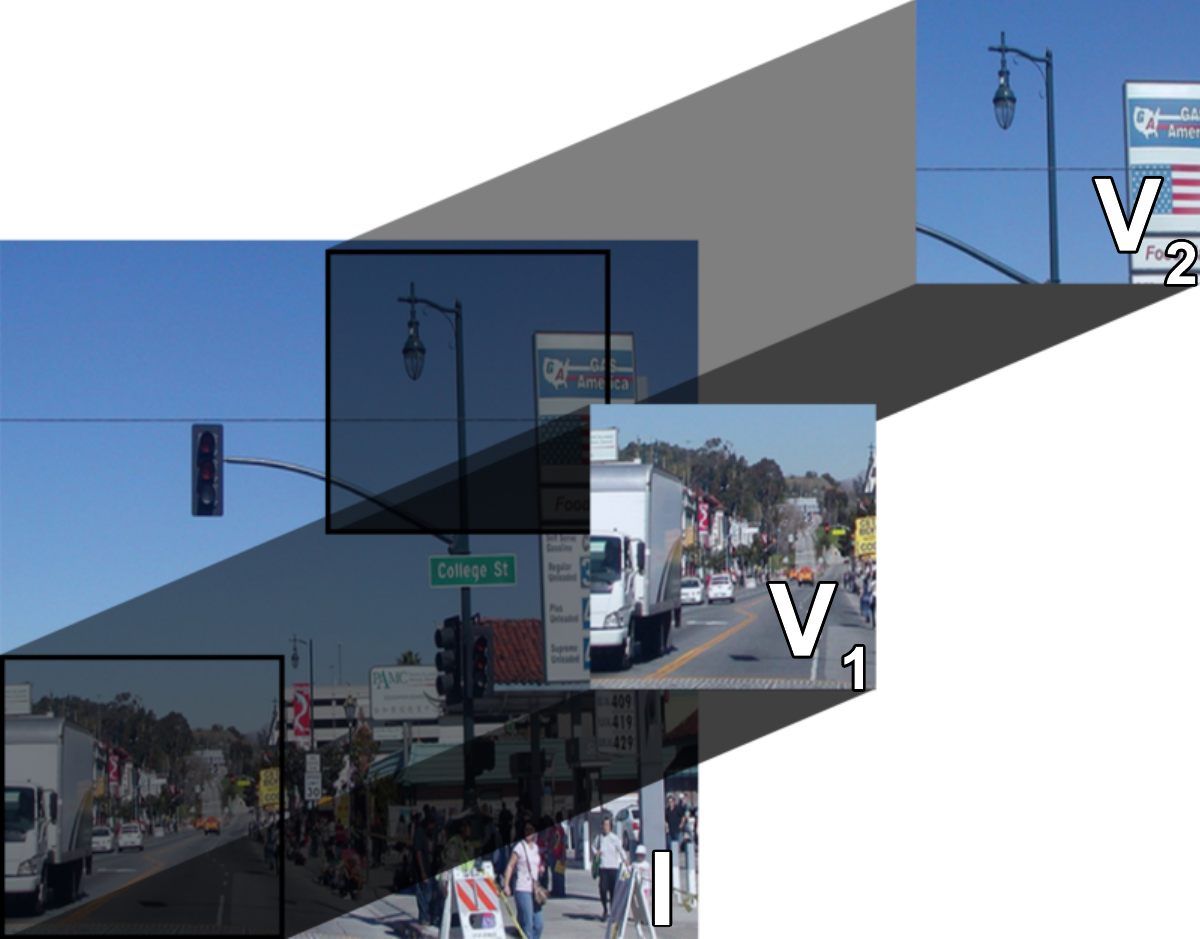}
    \caption{Random.}
    \label{subfig:random_views}
  \end{subfigure}
  \begin{subfigure}[t]{0.24\textwidth}
    \includegraphics[width=\textwidth]{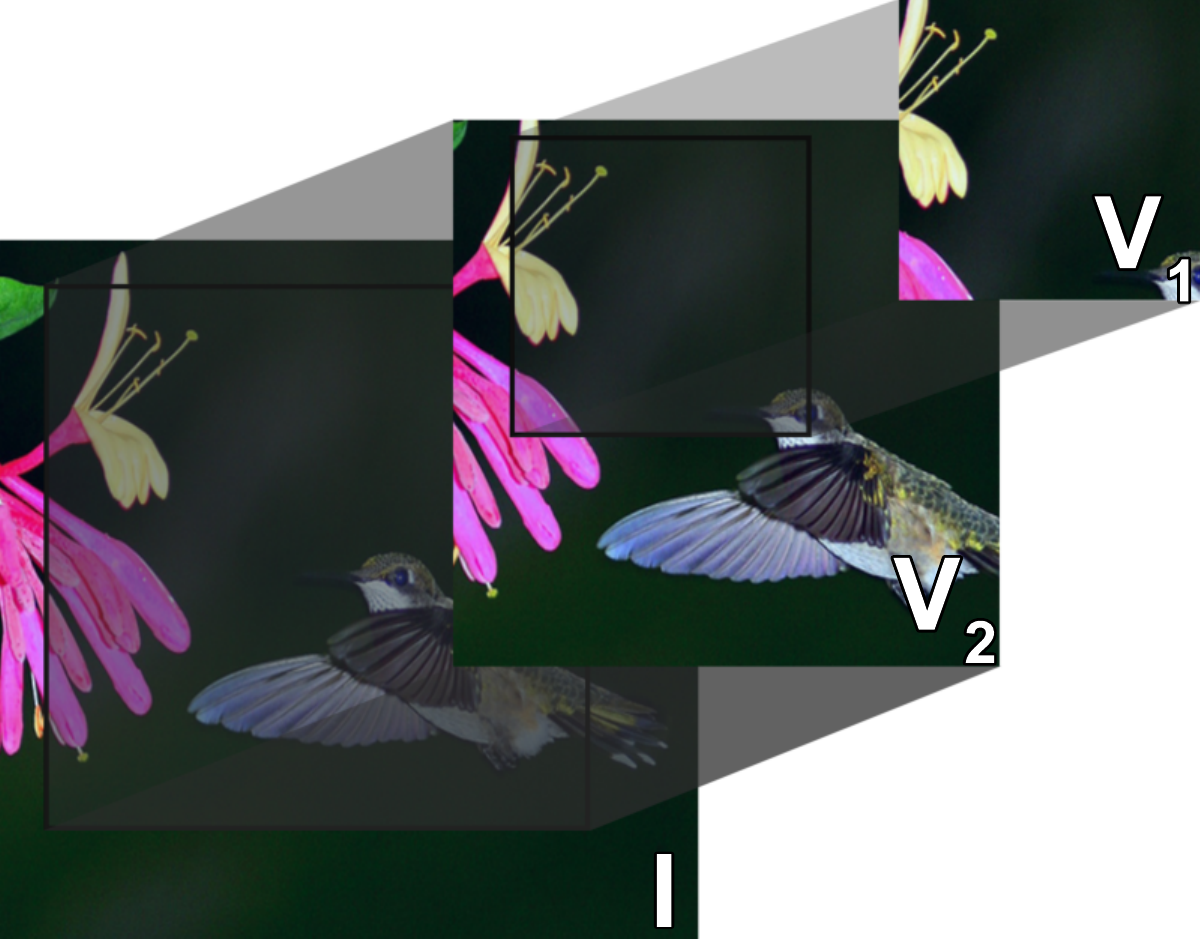}
    \caption{Local-to-Global.}
    \label{subfig:local_to_global}
  \end{subfigure}
  \begin{subfigure}[t]{0.24\textwidth}
    \includegraphics[width=\textwidth]{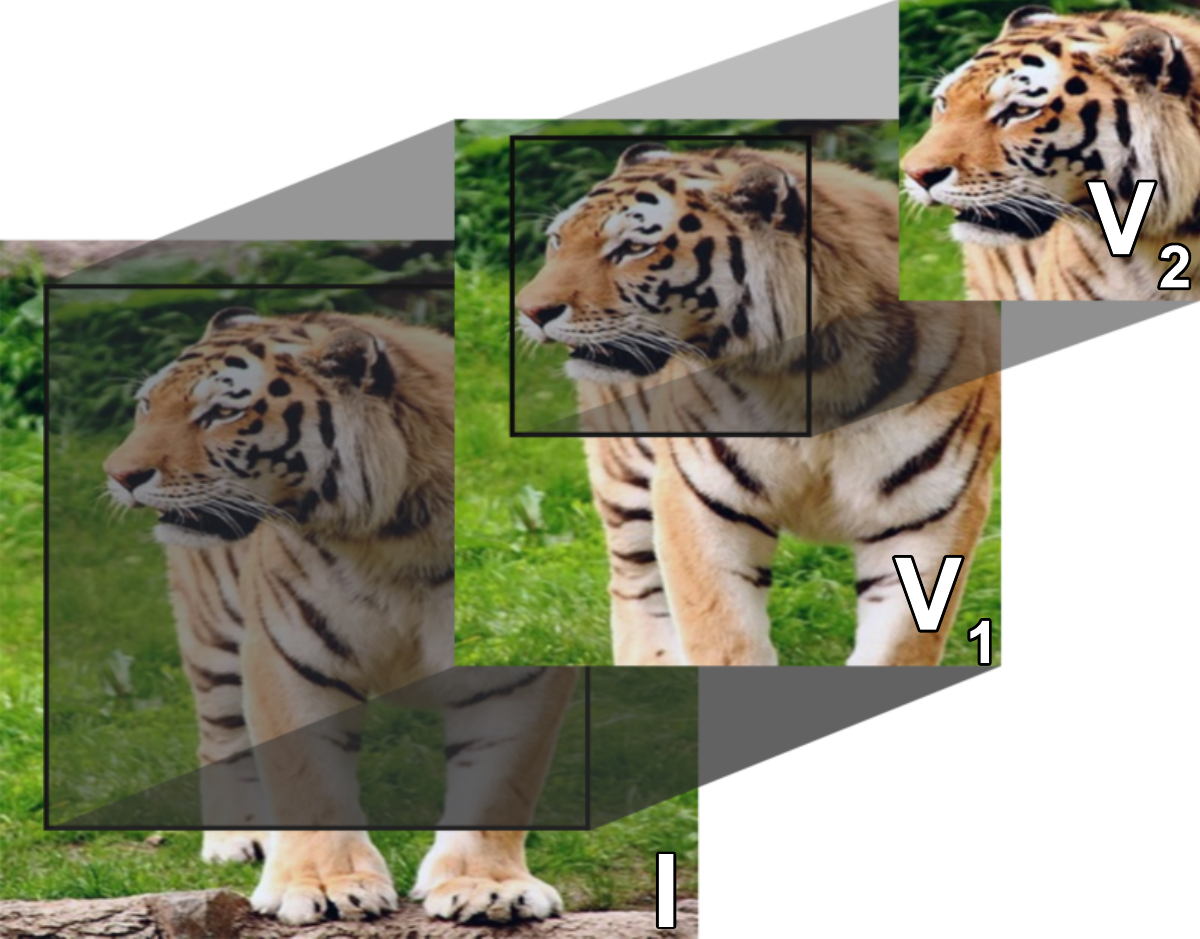}
    \caption{Global-to-Local.}
    \label{subfig:global_to_local}
  \end{subfigure}
    \caption{
     \textbf{Illustration of our four cropping strategies}. For a given input image $I$, we generate an unmasked view $V_1$ and a masked view $V_2$ following one of four different cropping strategies: (a) Same Views, where $V_1 = V_2$; (b) Random Views, where $V_1$ and $V_2$ are two independent random crops; (c) Local-to-Global, where $V_1$ is a random crop within $V_2$, and (d) Global-to-Local, where $V_2$ is a random crop within $V_1$.
    }
  \label{fig:global_local_adjacent_views}
\end{figure}

\begin{itemize}
    \item \textbf{Same Views.} 
    This setup corresponds to a direct adaptation of SiamMAE to images, in which the input image $I$ is cropped once and serves both as $V_1$ and $V_2$. An illustration is given in \Cref{subfig:same_views}.
    \item \textbf{Random Views.} For a given input image $I$, two independent random cropped views are generated for $V_1$ and $V_2$. 
    This setup poses a challenge, particularly when the views are adjacent, \ie, that there is minimal to no overlap between the two crops as illustrated in \Cref{subfig:random_views}.
    \item \textbf{Local-to-Global Views.}
    In this setup, the masked view $V_2$ is a random crop of the original image $I$, and the unmasked view $V_1$ is another random crop of the masked view $V_2$. An illustration is provided in \Cref{subfig:local_to_global}.
    \item \textbf{Global-to-Local Views.} Inversely, the unmasked view $V_1$ is a random crop of the original image $I$, and the masked view $V_2$ is another random crop of the unmasked view $V_1$. An illustration is provided in \Cref{subfig:global_to_local}.
\end{itemize}
Note that our experiments indicate that the Global-to-Local view strategy leads to the best performance.

\subsection{Patching}
\label{subsec:patching}
The two views $V_1$ and $V_2$ are patched following the original ViT~\cite{Dosovitskiy2021ViT}. Specifically, each view is converted into $N \times N$ patches that are fed into the encoder. 
Similar to SiamMAE, we augment the linear projections of these patches with positional embeddings~\cite{Vaswani2017Attention-arxiv}, and append a \texttt{[CLS]} token.

\subsection{Masking}
\label{subsec:masking}
Since both views are highly spatially redundant, a high masking ratio (above 75\%) is usually necessary to create a challenging pretext task and to achieve optimized performances with masked autoencoders~\cite{He2022Masked}. This is even more important in videos where both the spatial and temporal dimensions are highly redundant, requiring even higher masking ratios (90\%)~\cite{Tong2022VideoMAE, Wang2023VideoMAEV2, Feichtenhofer2022MAEST}.
SiamMAE~\cite{Gupta2023Siamese} employs a highly asymmetrical masking strategy, where the first frame is left completely visible while the second one is masked at $95\%$, which corresponds to $9$ visible patches out of the $196$ available when using a ViT/16~\cite{Dosovitskiy2021ViT}. Using such a high masking ratio encourages the model to propagate the visible patches from the first frame to the second one and to learn temporal correspondences through motion~\cite{Gupta2023Siamese}. However, employing a high masking ratio can make some examples ambiguous or may require additional knowledge beyond merely ``propagating'' patches from the unmasked view. For instance, if an object is only partially visible in the first view, while it is completely present (but masked) in the second one, the task becomes intractable if the model relies solely on the first view to reconstruct it. 
This prompts the model to learn a conceptual representation of the objects it encounters~\cite{He2022Masked}, enabling it to ``hallucinate'' what it partially sees when propagating past patches is either impossible or insufficiently informative.

Unlike previously introduced MAE methods, CropMAE does not need to learn any conceptual information about objects. Indeed, since our pretext task reconstructs a local view from a global one, there is no ambiguity as the local view is always present within the global view. Provided that the model \textbf{(i)} successfully identifies the location of the local view within the global view based on the visible patches and \textbf{(ii)} accurately determines the transformations required to reconstruct the local view from the global view, the task is directly tractable based on the inputs that the model receives without any prior conceptual knowledge. This naturally makes the pretext task significantly easier than in other MAE approaches such as MAE~\cite{He2022Masked}, VideoMAE~\cite{Tong2022VideoMAE}, or SiamMAE~\cite{Gupta2023Siamese}, where rich conceptual representations should be used to solve the task. For that reason, we employ an even higher masking ratio. More specifically, our method performs best with only a few visible patches, typically $1$ or $2$ out of $196$, which corresponds to a masking ratio between $98\%$ and $99\%$. Note that increasing the masking ratio from $95\%$ to $98.5\%$ decreases the number of visible patches by a factor of $4.5$, reducing them from $9$ to just two visible patches.

\subsection{Encoder and Decoder Architectures}
\label{subsec:architecture}
Following~\cite{Gupta2023Siamese}, we use a Siamese ViT~\cite{Dosovitskiy2021ViT} encoder to process our two views and a vanilla Transformer~\cite{Vaswani2017Attention-arxiv} composed of cross-attention and self-attention layers as our decoder. Specifically, our decoder alternates between self-attention, where tokens of the masked image attend to each other, and cross-attention layers, where the tokens of the masked image attend to tokens of the visible image.
We train the Siamese architecture by minimizing the L2 loss between the normalized~\cite{He2022Masked} pixel values of the view $V_2$ and the reconstruction $R$.
\section{Experiments}
\label{sec:Exp}

\subsection{Experimental setup}
\label{sec:Setup}

\mysection{Implementation details.} 
Following previous methods~\cite{Caron2021Emerging, Gupta2023Siamese, Tong2022VideoMAE}, we use the ViT-S/16 as encoder architecture~\cite{Dosovitskiy2021ViT} for most of our experiments and fair comparisons with respect to other methods in the field. For the decoder, we employ a 4-layer Transformer~\cite{Vaswani2017Attention-arxiv} with a dimension $d_{\text{model}}=256$, where each block comprises a cross-attention layer, a feed-forward layer (of dimension $d_{\text{ff}}=2048$), and a self-attention layer. GELU activation functions~\cite{Hendrycks2016Gaussian-arxiv} are utilized alongside a dropout rate of 10\%~\cite{Srivastava2014Dropout}.  We use the AdamW~\cite{Loshchilov2018Decoupled} optimizer and a base learning rate of $1.5e^{-4}$. The exhaustive list of hyper-parameters that we use can be found in the Appendix~\ref{sec:environment}.

\mysection{Baselines.}
We compare our method with several state-of-the-art methods including MAE-ST~\cite{Feichtenhofer2022MAEST}, MAE~\cite{He2022Masked}, VideoMAE~\cite{Tong2022VideoMAE}, and SiamMAE~\cite{Gupta2023Siamese}. To the best of our knowledge, no official open-source code is available for SiamMAE, so we reimplemented it to compare the evolution of our performance during training, using the exact same hyperparameters described in the SiamMAE paper (refer to the supplementary material). Our results are consistent with the ones reported in their paper~\cite{Gupta2023Siamese}. However, we train for $400$ epochs instead of $2000$ to save computational resources. Results for longer training can be found in the Appendix~\ref{subsec:longer_training_appendix}.

\mysection{Datasets.} 
We pre-train our models on Kinetics-400~\cite{Kay2017TheKinetics-arxiv} (K400), on ImageNet~\cite{Russakovsky2015ImageNet} (IN), or on a subset of ImageNet (IN Subset). 
IN Subset contains $239,787$ randomly selected images, which corresponds to the number of videos in K400, for fair comparison between methods trained on K400 and ImageNet. 
During pre-training, we randomly sample an image (or a frame on K400), which is then processed following our methodology described in \Cref{sec:Method}.

\mysection{Downstream tasks.}
We evaluate our method on three propagation downstream tasks: video object segmentation (DAVIS-2017~\cite{PontTuset2017Davis-arxiv}), human pose propagation (JHMDB~\cite{Jhuang2013Towards}) and semantic part propagation (VIP~\cite{Zhou2018Adaptive}). These propagation tasks are framed as a semi-supervised problem, where the first annotated frame is provided, and the model is expected to propagate the segmentation mask to subsequent frames.

\subsection{Results}
\label{sec:comparison_methods}

\begin{figure}[t!]
    \centering
    \includegraphics[width=\columnwidth]{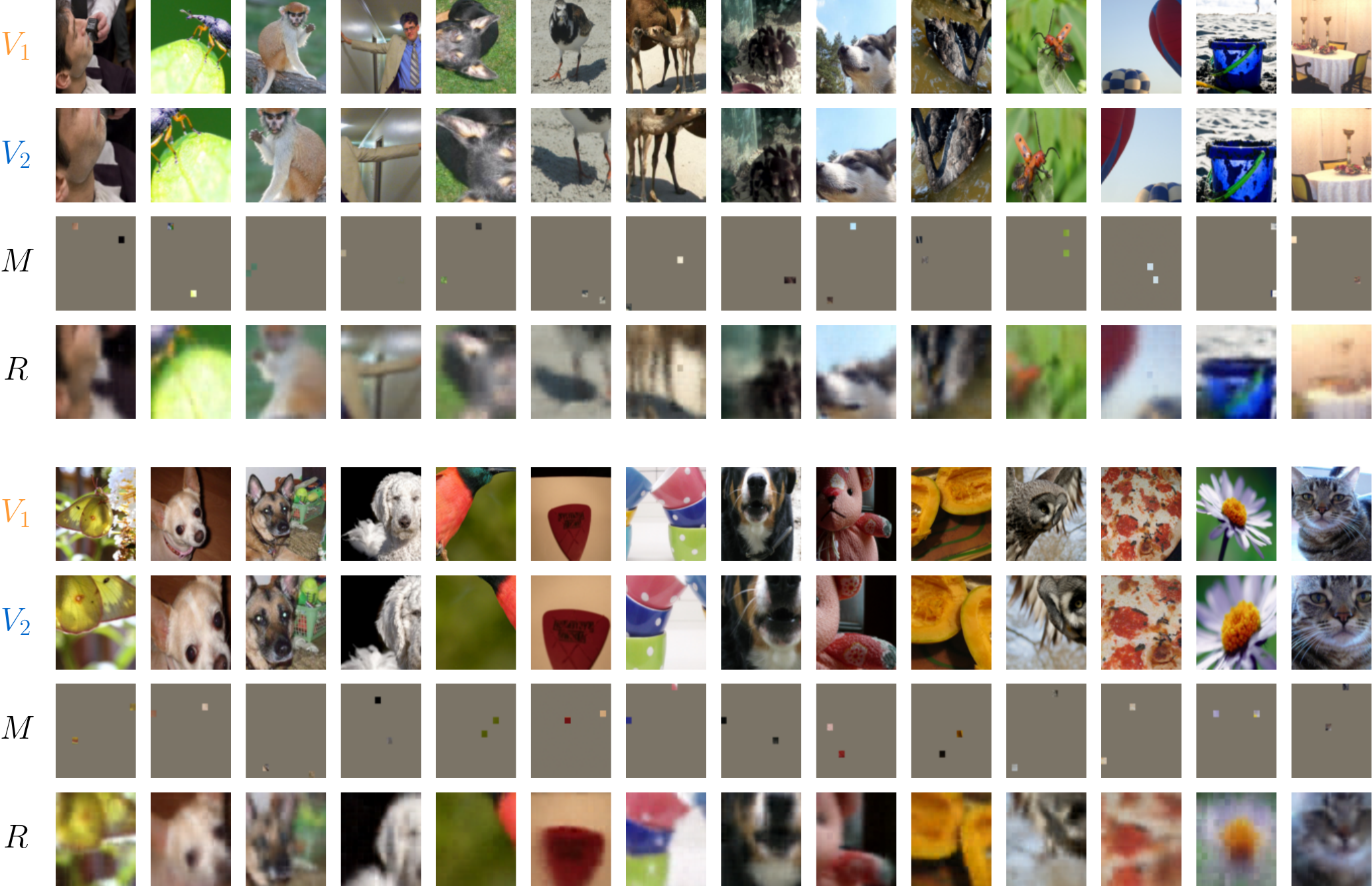}
    \caption{\textbf{Reconstructions of CropMAE.} We train CropMAE with a ViT-S/16 without normalizing pixel values and a masking ratio of 98.5\%. We visualize the reconstructions of some images from ImageNet. The images are displayed in the following order from top to bottom: Input Image ($V_1$), Random Resized Crop ($V_2$), Masked Image ($M$), and Reconstruction ($R$).}
    \label{fig:qualitative_results_reconstruction}
\end{figure}

\begin{figure}[t!]
    \centering
    \includegraphics[width=\columnwidth]{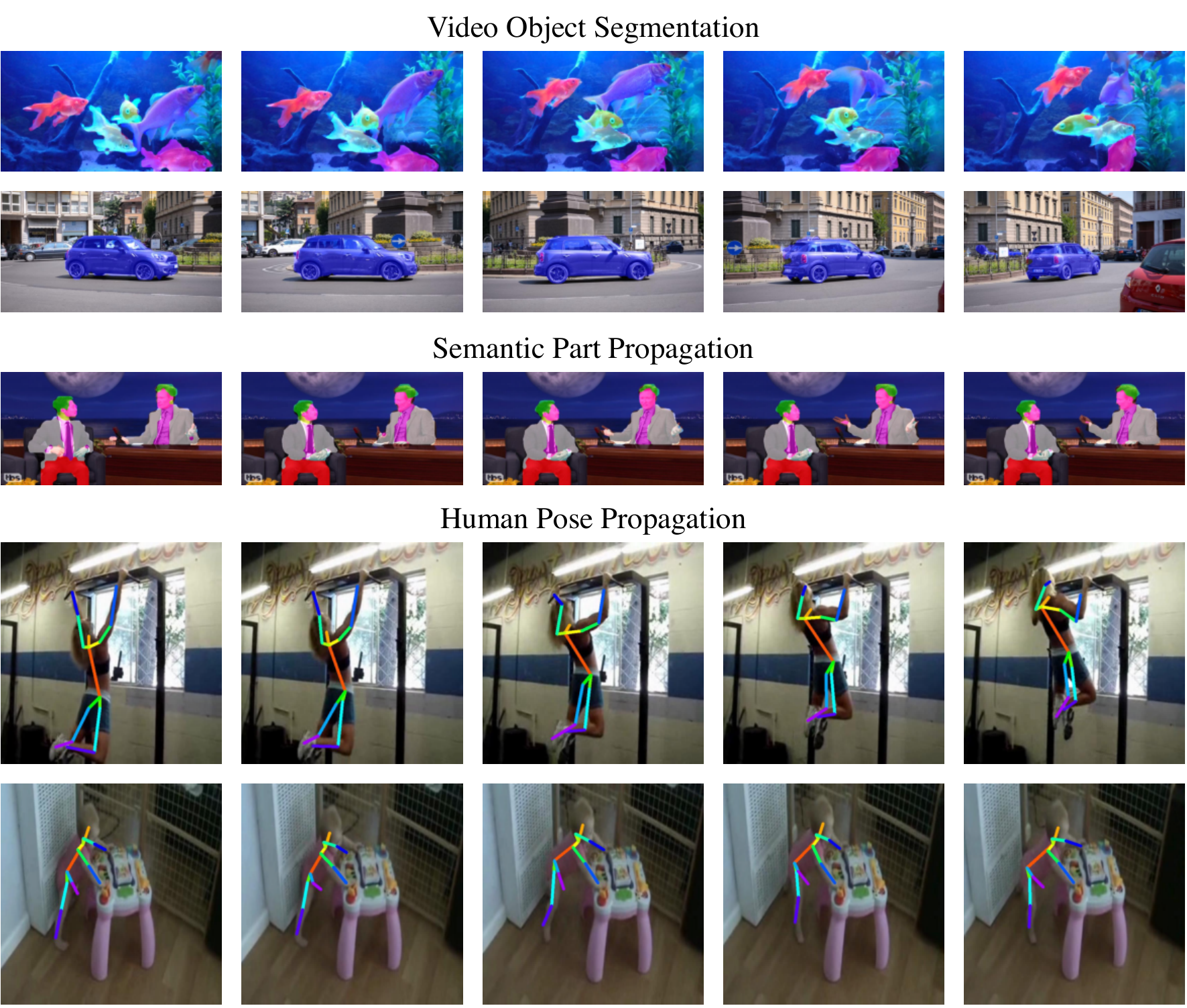}
    \caption{
    \textbf{Qualitative results.} We train CropMAE with a ViT-S/16 and qualitatively validate our results on three propagation downstream tasks: video object segmentation (DAVIS-2017~\cite{PontTuset2017Davis-arxiv}), semantic part propagation~\cite{Zhou2018Adaptive}, and human pose propagation (JHMDB~\cite{Jhuang2013Towards}).
    }
    \label{fig:qualitative_results_downstreams}
\end{figure}

We compare our method to previous works and present quantitative results in Table~\ref{tab:main_results}. We then provide some qualitative results of the reconstructed image and the downstream tasks respectively in \Cref{fig:qualitative_results_reconstruction,fig:qualitative_results_downstreams}. The first part of Table~\ref{tab:main_results} displays results as reported in their original papers, under optimal training conditions in terms of both training duration and data volume.
In the second part, we report the results achieved by our reproduced implementation of SiamMAE and CropMAE under our constrained training: either on K400 or on our ImageNet Subset, for a fixed duration of $400$ epochs, and for both ViT-S/16 and ViT-B/16. 

\begin{table}[t!]
\centering
\caption{
    \textbf{Comparison with prior work.} 
    We evaluate our method on three downstream tasks: video object segmentation (DAVIS-2017~\cite{PontTuset2017Davis-arxiv}), human pose propagation (JHMDB~\cite{Jhuang2013Towards}) and semantic part propagation (VIP~\cite{Zhou2018Adaptive}). 
    Specifically, we compare our method with other methods trained on 400 epochs, on K400~\cite{Kay2017TheKinetics-arxiv} or on our ImageNet~\cite{Deng2009ImageNet} Subset (IN Sub) for fair comparison. $\dagger$ refers to results reported in~\cite{Gupta2023Siamese}. $\ddagger$ refers to our implementation.
    }
\resizebox{\textwidth}{!}{%
   \begin{tabular}{l|ccc|ccc|c|cc}
&  &  & & \multicolumn{3}{c|}{DAVIS}  & \multicolumn{1}{c|}{VIP} & \multicolumn{2}{c}{JHMDB}\\
Method
& Backbone
& Dataset
& Epochs
& $\mathcal{J\&F}_m$ & $\mathcal{J}_m$ & $\mathcal{F}_m$ & mIoU   & PCK@0.1 & PCK@0.2   \\
\midrule
MAE-ST~\cite{Feichtenhofer2022MAEST} $\dagger$ & ViT-L/16 & K400 & 800 & 54.6 & 55.5 & 53.6 & 33.2 & 44.4 & 72.5 \\
MAE~\cite{He2022Masked} $\dagger$            & VIT-B/16 & IN & 1600 & 53.5 & 52.1 & 55.0 & 28.1 & 44.6 & 73.4 \\
VideoMAE~\cite{Tong2022VideoMAE} $\dagger$    & ViT-S/16 & K400 & 800 & 39.3 & 39.7 & 38.9 & 23.3 & 41.0 & 67.9 \\
SiamMAE~\cite{Gupta2023Siamese} $\dagger$    & ViT-S/16 & K400 & 2000 & \textbf{62.0} & \textbf{60.3} & \textbf{63.7} & \textbf{37.3} & \textbf{47.0} & \textbf{76.1} \\
\midrule
SiamMAE~\cite{Gupta2023Siamese} $\ddagger$    & ViT-S/16 & K400 & 400 & 57.9 & 56.0 & 60.0 & 33.2 & \textbf{46.1} & \textbf{74.0} \\
\textbf{CropMAE} (ours) & ViT-S/16 & K400 & 400 & 58.6 &  55.8 & 61.4 & \textbf{33.7} & 42.9 & 71.1 \\
\textbf{CropMAE} (ours)  & ViT-S/16 & IN Sub & 400 & \textbf{60.4} & \textbf{57.6} & \textbf{63.3} & 33.3 & 43.6 & 72.0 \\
\midrule
\textbf{CropMAE} (ours) & ViT-B/16 & IN Sub & 400 & 60.9 & 57.9 & 63.8 & 32.8 & 44.3 & 72.3 \\
\bottomrule
    \end{tabular}
}
    \label{tab:main_results}
\end{table}

When trained for $2{,}000$ epochs on K400, SiamMAE achieves state-of-the-art performances on the three downstream tasks, and outperforms previous MAE methods such as MAE-ST~\cite{Feichtenhofer2022MAEST}, MAE~\cite{He2022Masked} and VideoMAE~\cite{Tong2022VideoMAE}. However, considering a fixed budget of $400$ epochs, CropMAE achieves significantly better results than SiamMAE on DAVIS-2017~\cite{PontTuset2017Davis-arxiv}, both when trained on K400 and on our ImageNet Subset ($+0.7\%$ and $+2.5\%$ respectively). We believe that by explicitly transforming images through cropping, our pre-training method more quickly understands features useful for segmentation, such as object boundaries. On VIP~\cite{Zhou2018Adaptive}, CropMAE still performs better than SiamMAE, although by a smaller margin ($+0.1$ when trained on ImageNet, and $+0.5$ when trained on K400).
On JHMDB~\cite{Jhuang2013Towards}, CropMAE only outperforms VideoMAE. 
We explain these inferior performances by noting that SiamMAE uses two different frames, resulting in complex human pose modifications, which likely helps the network understand human motion and perform better on JHMDB. Conversely, our random crops do not mimic these transformations. Yet, they help the network learn object boundaries more explicitly, making it more suited for segmentation tasks such as DAVIS.

\subsection{Attention Maps}
\label{sec:attention_maps}

\begin{figure}[!ht]
    \centering
    \includegraphics[width=\columnwidth]{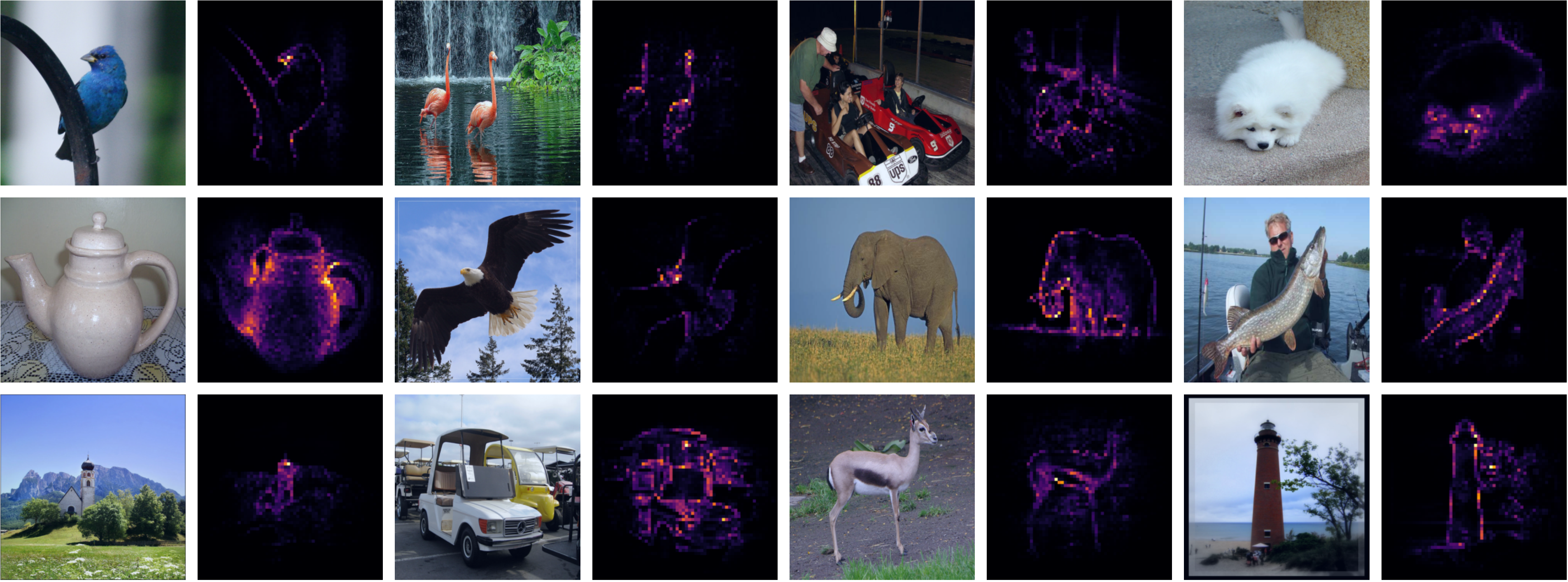}
    \caption{
    \textbf{Self-attention maps from CropMAE with a ViT-S/8 trained on our ImageNet subset.} We visualize the self-attention of the \texttt{[CLS]} token from a selected head in the last encoder layer of a ViT-S/8, which was trained on our ImageNet subset without using any supervision to learn this specific token. These self-attention maps reveal that our model can learn object boundaries without the need for prior motion information during pre-training.
    }
    \label{fig:vis_maps}
\end{figure}

In SiamMAE, Gupta~\etal~\cite{Gupta2023Siamese} argue that their model learns the concept of object boundaries through object motion in videos. To support this claim, they present attention maps extracted at some layers of their model, demonstrating that attention predominantly focuses on object boundaries. In a similar way, we train a ViT-S/8 with CropMAE on our ImageNet Subset and visualize the self-attention maps of the [CLS] token from a specific head of the last encoder layer. We show the results in \Cref{fig:vis_maps}. Our findings indicate that our model learns to identify object boundaries as well as SiamMAE without explicit motion (\ie, without relying on video frames). This implies that learning object boundaries is not solely attributable to the motion observed in videos; instead, it can also stem from the transformations and deformations operated on a single image. 
Hence, this phenomenon is present in both SiamMAE, where it happens naturally between two frames, and in CropMAE, where motion is artificially induced through random cropping. The main difference remains that CropMAE is trained on images instead of videos.

\subsection{Learning Speed}
\label{sec:training_speed}
 
\begin{figure}[!ht]
  \centering
  \includegraphics[width=0.9\textwidth]{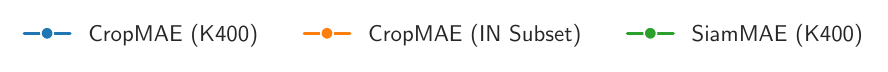}
    \centering
    \includegraphics[width=1.0\textwidth]{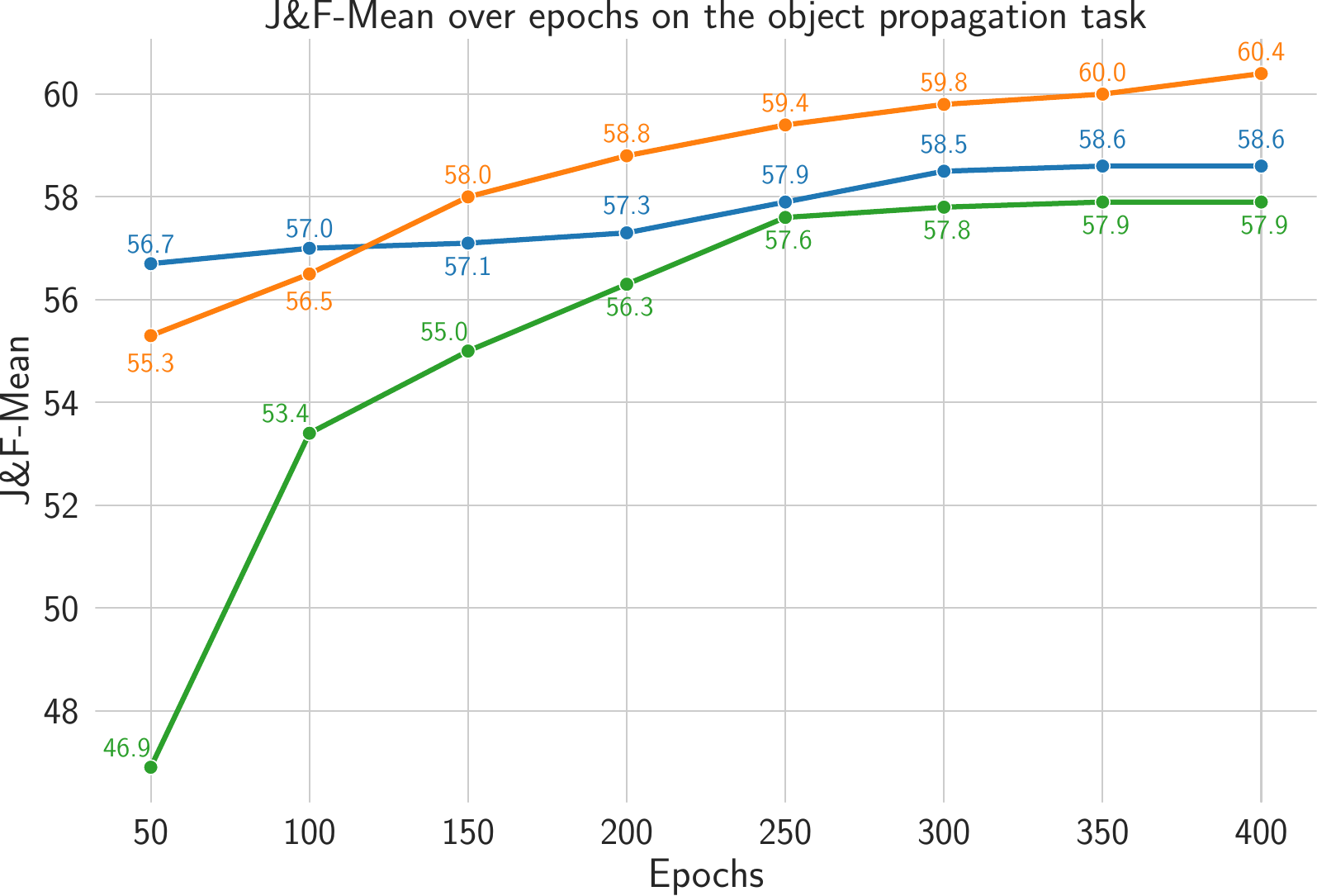}
    \caption{\textbf{Performances of CropMAE and SiamMAE on DAVIS during pre-training.} For a fixed number of $400$ epochs, CropMAE trains faster and consistently yields better results than SiamMAE~\cite{Gupta2023Siamese}, when trained on K400 frames or ImageNet Subset images.}
    \label{fig:training_speed_davis}
\end{figure}

We evaluate the evolution of the performances of CropMAE and SiamMAE. In particular, we compare SiamMAE trained on K400, CropMAE trained on K400, and CropMAE trained on ImageNet Subset, all for $400$ epochs. 
The performance on the DAVIS-2017 object propagation task~\cite{PontTuset2017Davis-arxiv} is reported every $50$ epochs in \Cref{fig:training_speed_davis}. Remarkably, our approach demonstrates superior performance when trained on the ImageNet Subset compared to training using K400 video frames. This improvement can be attributed to two main factors: \textbf{(i)} the greater diversity of the ImageNet dataset, containing a broader spectrum of objects, and \textbf{(ii)} its focus on currated object-centric images, which likely results in more relevant crops and reconstruction tasks. In contrast, random cropping in K400 frequently yields images without any objects, diminishing the effectiveness of the learning process.

Our approach demonstrates significantly faster learning than SiamMAE. In particular, our method achieves a $\mathcal{J}$\&$\mathcal{F}_m$ value of $58.0$ after only $150$ epochs on our ImageNet Subset and $250$ epochs on K400. In contrast, SiamMAE reaches the same performance level after $350$ epochs. We attribute this trend to our pretext task, which does not require any conceptual knowledge to be completely tractable and uses object transformations much more explicitly than SiamMAE, leading to faster propagation comprehension. In contrast, SiamMAE must learn the concept of motion and understand object transformations more implicitly between two frames through more complex perturbations such as occlusions and viewpoint changes.

\subsection{Training time}
We compare the training times of CropMAE and SiamMAE. On the one hand, CropMAE uses an extremely high masking ratio, and only needs a single frame of a video clip to train, or even a standalone image. On the other hand, SiamMAE uses a lower masking ratio and needs two different frames to work.  Both these factors significantly impact the training time, as seeking distant frames may require decoding a larger portion of the video, and the number of operations performed by the attention layers increases quadratically with the number of visible patches~\cite{He2022Masked}. We measure the total time taken by both approaches to train and report our results in Table~\ref{tab:training_time}. As it can be seen, CropMAE trains almost $30\%$ faster than SiamMAE on K400 for a fixed computational budget, thanks to its use of fewer patches and frames. When pre-training on images (\ie, on the IN Subset), which are significantly faster to decode, CropMAE achieves a tremendous speed-up of 2380\% on our hardware while also reaching better performances.

\begin{table}[t!]
    \centering
    \caption{
    \textbf{Speedup of CropMAE compared to SiamMAE.} 
    We train both methods for $400$ epochs on K400, and on ImageNet Sub for CropMAE, and report the speedups observed on the whole training process.
    }
    \begin{tabular}{@{}cccccc@{}}
        \toprule
        \textbf{Method} & Dataset & Number of images &  Mask Ratio & GFLOPS & Speedup\\ 
        \midrule
        SiamMAE & K400 & 2 & 95\% & 5.8 & $\times$1.0\\
        CropMAE & K400 & 1 & 98.5\% & 5.6 & $\times$1.29\\
        CropMAE & IN Subset & 1 & 98.5\% & 5.6 & $\times$23.8\\
        \bottomrule
    \end{tabular}
    \label{tab:training_time}
\end{table}

\subsection{Ablation Studies}
\label{sec:Ablation}

We perform several ablation studies on the different components of CropMAE and report the results in Table~\ref{tab:ablations_1}. Unless stated otherwise, we use the default parameters presented in the Appendix~\ref{sec:environment}. Specifically, we train CropMAE on our ImageNet subset for $400$ epochs and report the results obtained on the DAVIS-2017~\cite{PontTuset2017Davis-arxiv} object propagation task.

\begin{table*}[t!]
    \centering
    \caption{
    \textbf{Ablation Study.} 
    We analyze the different components of our method to understand their impact on the downstream performance. 
    We use a ViT-S/16~\cite{Dosovitskiy2021ViT} with the default configuration, as presented in Section~\ref{sec:Setup}, and report the results obtained on the DAVIS-2017~\cite{PontTuset2017Davis-arxiv} validation set.
    }
    \subfloat[
    \textbf{Crop Strategy.} A simple extension of SiamMAE to images does not work. Reconstructing the local view from the global view works best for CropMAE.
    \label{fig:crop_strategy_results}
    ]{
    \centering
    \begin{minipage}{0.50\linewidth}
    \begin{center}
    
\begin{tabular}{@{}cccc@{}}
\toprule
\textbf{Crop Strategy} & $\mathcal{J}$\&$\mathcal{F}_m$ & $\mathcal{J}_m$ & $\mathcal{F}_m$ \\ 
\midrule
Same Views & 36.6 & 35.8  & 37.5  \\
Random Views & 60.0 & 57.2 & 62.8 \\
Local-to-Global & 55.9  & 53.8   & 58.0  \\
\baseline{\textbf{Global-to-Local}} & \baseline{\textbf{60.4}}  & \baseline{\textbf{57.6}}  & \baseline{\textbf{63.3}}  \\
\label{tab:crop_strategy}
\end{tabular}
    \end{center}
    \end{minipage}
    }
    \hspace{1em}
    \subfloat[
    \textbf{Mask Ratio and number of visible patches.} Our method works best when an extremely large portion of the patches is masked. 
     \label{fig:mask_ratio_results}
    ]{
    \centering
    \begin{minipage}{0.40\linewidth}
    \begin{center}
    \begin{tabular}{@{}cccc@{}}
\toprule
\textbf{Mask Ratio} &  $\mathcal{J}$\&$\mathcal{F}_m$ & $\mathcal{J}_m$ & $\mathcal{F}_m$ \\ 
\midrule
0.75 (49) & 45.3  & 44.3  & 46.3  \\
0.90 (19)& 47.1  & 46.1 & 48.0  \\
0.95 (9)& 51.2  & 49.9  & 52.4  \\
\baseline{\textbf{0.985 (2)}} & \baseline{\textbf{60.4}}  & \baseline{\textbf{57.6}}  & \baseline{\textbf{63.3}}   \\
0.99 (1) & 58.6  & 55.9   & 61.5  \\
\label{tab:masking}
\end{tabular} 
    \end{center}
    \end{minipage}
    }
    \\[2mm]
    \subfloat[
    \textbf{Decoder Depth.} Our method works best with a small depth.
    \label{fig:decoder_depth}
    ]{
    \centering
    \begin{minipage}{0.45\linewidth}
    \begin{center}
    \begin{tabular}{@{}cccc@{}}
\toprule
\textbf{Decoder Depth} &  $\mathcal{J}$\&$\mathcal{F}_m$ & $\mathcal{J}_m$ & $\mathcal{F}_m$ \\ 
\midrule
2 & 59.1  & 56.7  & 61.6  \\
\baseline{\textbf{4}} & \baseline{\textbf{60.4}}  & \baseline{\textbf{57.6}}  & \baseline{\textbf{63.3}}   \\
8 & 57.0  & 54.5  & 59.4  \\
\label{tab:decoder_arch}
\end{tabular}
    \end{center}
    \end{minipage}
    }
    \hspace{1em}
    \subfloat[
    \textbf{Decoder Embedding Dimension.} Our method works best with a small decoder embedding dimension.
    \label{fig:decoder_embed_dim}
    ]{
    \centering
    \begin{minipage}{0.45\linewidth}
    \begin{center}
    \begin{tabular}{@{}cccc@{}}
\toprule
\textbf{Decoder Embed Dim} &  $\mathcal{J}$\&$\mathcal{F}_m$ & $\mathcal{J}_m$ & $\mathcal{F}_m$ \\ 
\midrule
128 & 58.5  & 56.0  & 61.0  \\
\baseline{\textbf{256}} & \baseline{\textbf{60.4}}  & \baseline{\textbf{57.6}}  & \baseline{\textbf{63.3}}   \\
384 & 59.0  & 56.3  & 61.7  \\
\label{tab:dec_embed_dim}
\end{tabular} 
    \end{center}
    \end{minipage}
    
    }
    \hspace{1em}
    \subfloat[
    \textbf{Data Augmentations.} Our method works best with horizontal flips randomly applied on both random crops.
    \label{fig:data_augmentations}
    ]{
    \centering
    \begin{minipage}{0.45\linewidth}
    \begin{center}
    
\begin{tabular}{@{}cccc@{}}
\toprule
\textbf{Augmentation} & $\mathcal{J}$\&$\mathcal{F}_m$ & $\mathcal{J}_m$ & $\mathcal{F}_m$ \\ 
\midrule
Color Jitter & 56.2 &  53.1 & 59.2  \\
Gaussian Blur & 59.6 & 56.7 & 62.4  \\
None & 60.3 & 57.4 & 63.2 \\
\baseline{\textbf{Horizontal flip}} & \baseline{\textbf{60.4}} & \baseline{\textbf{57.6}} & \baseline{\textbf{63.3}}
\label{tab:crop_strategy}
\end{tabular}
    \end{center}
    \end{minipage}
}
    \label{tab:ablations_1} 
\end{table*}

\mysection{Cropping Strategy.}
\label{sec:Crop_Strategy}

We study the effect on performance of different cropping strategies in Table~\ref{fig:crop_strategy_results}. 
We can see that reconstructing the same views (\Cref{subfig:same_views}) yields very poor performances ($36.6$), suggesting that the model failed to learn any propagation capabilities. 
Reconstructing the Local-to-Global view (\Cref{subfig:local_to_global}) results in significantly improved performance ($55.9$).
The Random Views (\Cref{subfig:random_views}) and Global-to-Local (\Cref{subfig:global_to_local}) approaches achieve the highest scores ($60.0$ and $60.4$, respectively). 
Interestingly, these setups are the only ones enabling a completely tractable task without any prior knowledge, meaning the reconstruction can solely rely on the unmasked image. In fact, tractability is \textit{sometimes} guaranteed in the random setting, while it is \textit{always} true for the Global-to-Local approach, which likely explains its superior performance.

\mysection{Masking Ratio.}
\label{sec:masking_ratio}

We examine the importance of the masking ratio in Table~\ref{fig:mask_ratio_results}. 
Our method exhibits suboptimal performance at a 75\% masking ratio, despite this being the preferred choice for the traditional image MAE framework~\cite{He2022Masked}. 
Similarly, it underperforms at the 90\% ratio used in video frameworks~\cite{Tong2022VideoMAE, Wang2023VideoMAEV2, Feichtenhofer2022MAEST}. 
We can see an improvement with a masking ratio of 95\%, as adopted in SiamMAE~\cite{Gupta2023Siamese}, but the optimal results are reached with a visibility reduced to merely a few patches, \ie, two ($60.4$) or one ($58.6$), equivalent of masking ratios of $98.5\%$ and $99\%$, respectively. We attribute this trend to the fact that our pretext task is simpler than those used in other frameworks as it does not require any conceptual knowledge and can be fully achieved with the help of the visible image, thus requiring an extremely high masking ratio to be challenging.

\mysection{Decoder Architecture.}
\label{sec:decoder_architecture}

Next, we study different decoder architectures, specifically their depth and embedding dimension. We report our results in Tables~\ref{fig:decoder_depth} and~\ref{fig:decoder_embed_dim}. Similarly to other MAE works~\cite{He2022Masked, Yao2020Video}, we found that the optimal decoder (256-d, 4 blocks) is smaller than the encoder (384-d, 12 blocks).

\mysection{Data Augmentations.}
\label{sec:data_augmentations}

We evaluate our method with additional data augmentations commonly used in contrastive learning~\cite{Grill2020Bootstrap, Chen2020Simple} and present our results in Table~\ref{fig:data_augmentations}. 
Similar to SiamMAE~\cite{Gupta2023Siamese}, we observe that using color jitter significantly reduces performance. 
The use of Gaussian blur also leads to a decline in performance but to a lesser extent. 
When we do not apply the random horizontal flip, we observe a minimal drop in performance.

\section{Conclusion}
\label{sec:Conclusion}

In this work, we introduce CropMAE, a self-supervised method for quickly learning rich features for video propagation tasks by reconstructing a crop of an image that has been masked at an extremely high proportion (over 98.5\%). We empirically demonstrate that our method can learn useful features for video downstream tasks without requiring explicit video motion. These features can be learned from still images, resulting in even richer information. Thanks to our tractable pretext task, our method trains faster than existing methods and is applicable to both video frames and still images. Finally, we show on-par performances with state-of-the-art methods for three video propagation downstream tasks.

\mysection{Limitations and future work.} Despite being designed to work with small quantities of data and facilitate fast training, we believe the scalability of our method warrants further investigation. 
This includes both model scalability (\ie, patch size and ViT size) and data scalability (\ie, the amount of data available and the differences between images and video frames). 
More effort should be directed towards understanding the unique contributions of video frames instead of still images, especially concerning scalability, and determining their necessity to develop rich and robust representations.

\section*{Acknowledgements}

A. Cioppa is funded by the F.R.S.-FNRS. 
The research reported in this publication was supported by funding from KAUST Center of Excellence on GenAI, under award number 5940, and the SDAIA-KAUST Center of Excellence in Data Science and Artificial Intelligence. 
The present research benefited from computational resources made available on Lucia, the Tier-1 supercomputer of the Walloon Region, infrastructure funded by the Walloon Region under the grant agreement n°1910247. 
We acknowledge EuroCC Belgium for awarding this project access to the LUMI supercomputer, owned by the EuroHPC Joint Undertaking, hosted by CSC (Finland) and the LUMI consortium.

%
%
\bibliographystyle{splncs04}
\bibliography{bib/abbreviation-short,bib/action,bib/dataset,bib/labo,bib/learning-supervision,bib/learning,bib/new-refs.bib,bib/self-supervised,bib/vision}

\clearpage

\section{Appendix}
\label{sec:supplementary}

\subsection{Evaluation environment}
\label{sec:environment}
Our implementation of CropMAE and SiamMAE are based on the MAE Pytorch open-source implementation\footnote{\url{https://github.com/facebookresearch/mae}}. Unless specified otherwise in Table \ref{tab:hyperparameters_cropmae_siammae}, we use the default parameters specified in the original paper~\cite{He2022Masked}. For the evaluation on the downstream tasks, we use the parameters presented in Table \ref{tab:params_downstreams}. Our experiments were performed on $4\times4$ NVIDIA A100 40GB and on $4\times$ AMD EPYC 7513 32-core. Video decoding on K400 was performed on CPU.

\begin{table}[h]
\caption{\textbf{Hyperparameters of CropMAE and SiamMAE.} Comparison of hyperparameters used for CropMAE, both on ImageNet~\cite{Deng2009ImageNet} and K400~\cite{Kay2017TheKinetics-arxiv}, and SiamMAE on K400~\cite{Kay2017TheKinetics-arxiv}. The same parameters were used for both methods when possible, and the original parameters of SiamMAE were used.}
\centering
\begin{tabular}{l|l|l}
\hline
\textbf{Config} & \textbf{CropMAE} & \textbf{SiamMAE~\cite{Gupta2023Siamese}} \\ \hline
Optimizer & AdamW~\cite{Loshchilov2018Decoupled} & AdamW~\cite{Loshchilov2018Decoupled} \\
Optimizer Momentum & $\beta_1, \beta_2=0.9, 0.95$~\cite{Chen2020Generative} & $\beta_1, \beta_2=0.9, 0.95$~\cite{Chen2020Generative} \\
Weight Decay & 0.05 & 0.05 \\
Learning Rate & 1.5e-4 & 1.5e-4 \\
Mask Ratio & 0.985 & 0.95 \\
Learning Rate Schedule & Cosine Decay~\cite{Loshchilov2017SGDR} & Cosine Decay~\cite{Loshchilov2017SGDR} \\
Warmup Epochs~\cite{Goyal2017Accurate-arxiv} & 10 & 10 \\
Epochs & 400 & 400 \\
Repeated Sampling~\cite{Hoffer2020Augment} & 1 (IN), 2 (K400) & 2 \\
Augmentation $V_1$ & Hflip (p=0.5), Crop $[a, c]$ \hphantom & Hflip (p=0.5), Crop $[0.5, 1]$ \\
Augmentation $V_2$ & Hflip (p=0.5), Crop $[b, d]$ & - \\
Effective Batch Size & 2048 & 2048 \\
Frame Sampling Gap & - & $[4, 48]$ \\ 
Min Aspect Ratio & 3/4 ($V_1$ \& $V_2$) & 3/4 \\ 
Max Aspect Ratio & 4/3  ($V_1$ \& $V_2$) & 4/3 \\ 
Min Area $V_1$ ($a$) & 0.10 (IN), 0.50 (K400) & - \\
Min Area $V_2$ ($b$) & 0.30 & - \\
Max Area $V_1$ ($c$) & 1.0 & - \\
Max Area $V_2$ ($d$) & 0.60  & - \\
\end{tabular}
\label{tab:hyperparameters_cropmae_siammae}
\end{table}

\begin{table}[h]
\caption{\textbf{Parameters used for the downstream tasks.}}
\centering
\begin{tabular}{l|l|l|l}
\hline
\textbf{Config} & \textbf{DAVIS-2017}~\cite{PontTuset2017Davis-arxiv} & \textbf{VIP}~\cite{Zhou2018Adaptive} & \textbf{JHMDB}~\cite{Jhuang2013Towards} \\ \hline
Top-k & 7 & 10 & 7 \\
Queue Length & 20 & 20 & 20 \\
Neighborhood Size & 20 & 20 & 20 \\
\end{tabular}

\label{tab:params_downstreams}
\end{table}

\subsection{Longer training}
\label{subsec:longer_training_appendix}

\begin{figure}[!t]
  \centering
  \includegraphics[width=0.9\textwidth]{figures/compressed/legend.pdf}
    \centering
    \includegraphics[width=1.0\textwidth]{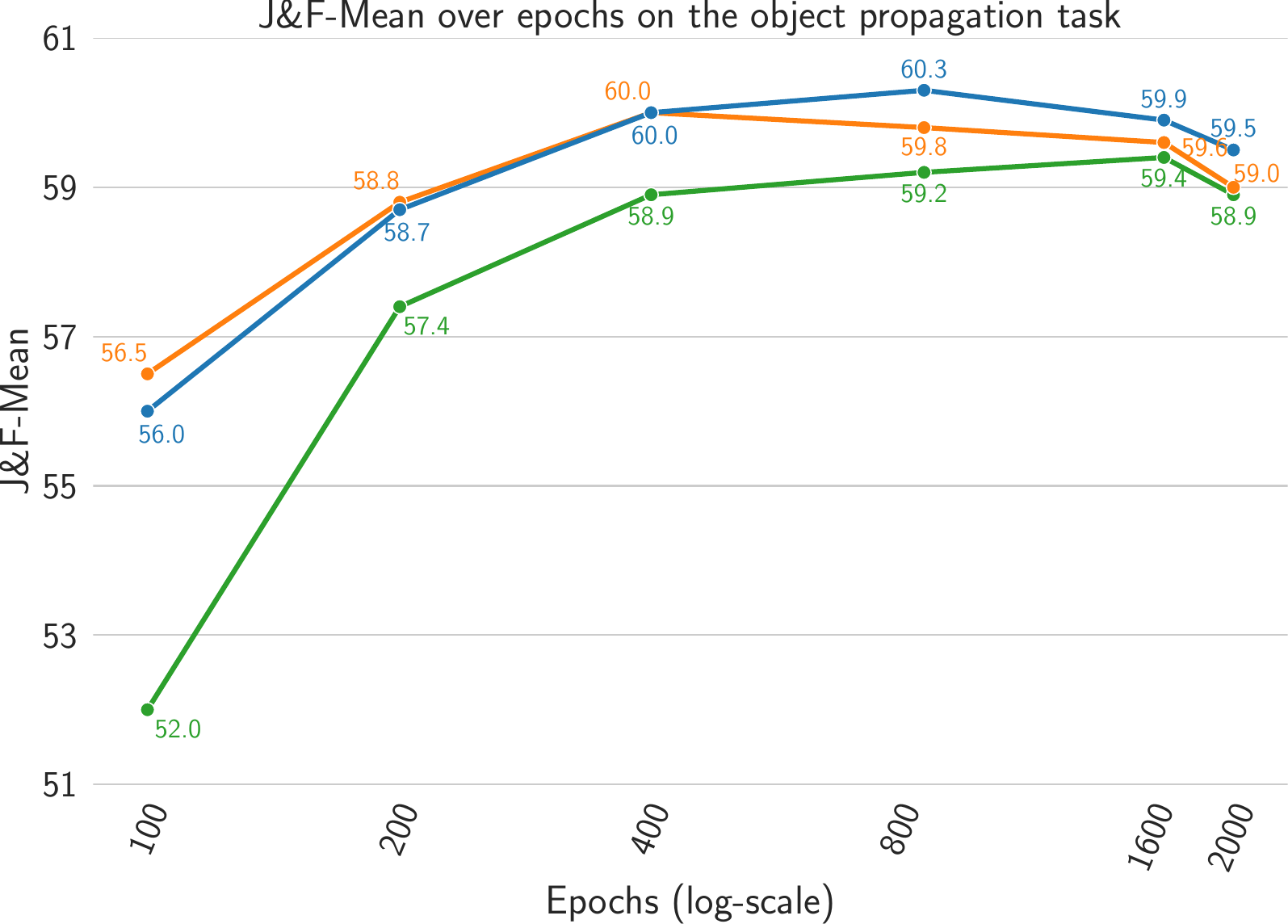}
    \caption{
    \textbf{Performances of CropMAE and SiamMAE on DAVIS during pre-training.} 
    For a fixed number of $2{,}000$ epochs, CropMAE trains faster and consistently yields better results than SiamMAE~\cite{Gupta2023Siamese}, when trained on K400 frames or ImageNet Subset images.}
    \label{fig:training_speed_davis_2k}
\end{figure}

We ran experiments for $2{,}000$ epochs with our different setups: SiamMAE trained on K400, CropMAE trained on K400, and CropMAE trained on ImageNet.
The results are presented in \Cref{fig:training_speed_davis_2k}. 

Overall, our approach demonstrates consistent superior performance compared to SiamMAE for both video (K400) and image (ImageNet) training. Our approach demonstrates significantly faster learning than SiamMAE. In particular, our method achieves a $\mathcal{J}$\&$\mathcal{F}_m$ value of $56.5$ after only $100$ epochs on our ImageNet Subset, whereas SiamMAE only achieves a value of $52.0$ at this stage. At $400$ epochs, our method reaches a $\mathcal{J}$\&$\mathcal{F}_m$ value of $60.0$ for both video and image training, while SiamMAE has a value of $58.9$. Even though the peak value ($59.4$) of SiamMAE is achieved later during the pre-training compared to our method, SiamMAE is not able to reach the performance we obtain. We attribute this trend to our pretext task, which does not require any conceptual knowledge to be completely tractable and uses object transformations much more explicitly than SiamMAE, leading to faster propagation comprehension. In contrast, SiamMAE must learn the concept of motion and understand object transformations more implicitly between two frames through more complex perturbations such as occlusions and viewpoint changes. Finally, we can see that none of the three methods is really able to scale well with very long pre-training, which is a behavior already depicted in \cite{Jiang2023Concatenated-arxiv}.


\end{document}